\documentclass[conference]{IEEEtran}
\IEEEoverridecommandlockouts

\usepackage{booktabs}
\usepackage{multirow}
\usepackage{cite}
\usepackage{amsmath,amssymb,amsfonts}
\usepackage{algorithmic}
\usepackage{graphicx}
\usepackage{textcomp}
\usepackage{xcolor}
\usepackage{float}

\def\BibTeX{{\rm B\kern-.05em{\sc i\kern-.025em b}\kern-.08em
    T\kern-.1667em\lower.7ex\hbox{E}\kern-.125emX}}
\begin{document}

\title{Structured Phonological Representations for Audio–Articulatory rtMRI Speech Classification
{\footnotesize \textsuperscript{}}
\thanks{}
}
\author{\IEEEauthorblockN{\textsuperscript{} Anonymous submission}
\IEEEauthorblockA{\textit{} \\
\textit{}\\
\\
}
\and
\IEEEauthorblockN{\textsuperscript{} Anonymous Institution}
\IEEEauthorblockA{\textit{} \\
\textit{}\\
\\
}
}

\author{
\IEEEauthorblockN{Abner Hernandez}
\IEEEauthorblockA{
\textit{Pattern Recognition Lab} \\
\textit{Friedrich-Alexander-Universität Erlangen-Nürnberg (FAU)}\\
Erlangen, Germany \\
abner.hernandez@fau.de
}
\and
\IEEEauthorblockN{Tom\'as Arias-Vergara}
\IEEEauthorblockA{
\textit{Pattern Recognition Lab} \\
\textit{Friedrich-Alexander-Universität Erlangen-Nürnberg (FAU)}\\
Erlangen, Germany \\
\textit{Department of Electronic Engineering}\\
\textit{Universidad de Antioquia}\\
Medell\'in, Colombia
}
\and
\IEEEauthorblockN{Daiqi Liu}
\IEEEauthorblockA{
\textit{Pattern Recognition Lab} \\
\textit{Friedrich-Alexander-Universität Erlangen-Nürnberg (FAU)}\\
Erlangen, Germany
}
\and
\IEEEauthorblockN{Andreas Maier}
\IEEEauthorblockA{
\textit{Pattern Recognition Lab} \\
\textit{Friedrich-Alexander-Universität Erlangen-Nürnberg (FAU)}\\
Erlangen, Germany
}
\and
\IEEEauthorblockN{Paula Andrea P\'erez-Toro}
\IEEEauthorblockA{
\textit{Pattern Recognition Lab} \\
\textit{Friedrich-Alexander-Universität Erlangen-Nürnberg (FAU)}\\
Erlangen, Germany \\
\textit{Department of Electronic Engineering}\\
\textit{Universidad de Antioquia}\\
Medell\'in, Colombia
}
}

\maketitle

\begin{abstract}
Real-time MRI makes it possible to observe vocal-tract articulation during speech, but mapping these articulatory patterns to phonetic and phonological categories remains challenging. We investigate whether PhonoQ, an audio-based model trained to recognize structured phonological features, provides useful information for audio--articulatory modeling. Specifically, we extract representations from PhonoQ's Conformer module, whose training is shaped by supervision for manner, place, voicing, and vowel features. Using articulatory contours with synchronized audio-derived features, we compare WavLM-large and HuBERT-large baselines with models that incorporate PhonoQ-derived representations. Across unseen-speech and unseen-subject settings, these features improve macro-F1 for phonological targets including manner, place, voicing, vowel height, and vowel backness, and also improve fine-grained 39-phoneme classification. In a contour-only inference setting, audio-derived teacher supervision yields modest but consistent gains over contour-only training, indicating that phonological information from synchronized audio can be partially transferred to articulatory models. Finally, posterior analyses show interpretable surface-sensitive patterns consistent with flapping-like \mbox{/t/} realizations, \mbox{/t/--/r/} retraction or affrication, and nasal place assimilation.
\end{abstract}


\section{Introduction} 
\label{sec:introduction}

Speech sounds arise from rapid, coordinated vocal-tract movements, yet most computational speech models learn primarily from the acoustic signal. Real-time magnetic resonance imaging (rtMRI) offers direct visual access to this articulatory process by capturing the motion of the tongue, lips, velum, and pharyngeal walls during continuous speech, making it valuable for linking speech production to phonetic and phonological structure~\cite{ziegler2019motor,richmond2012ultrax,scott2014speech}. Mapping articulatory observations to linguistic categories, however, remains challenging. Midsagittal contours encode vocal-tract shape, but not every phonemic contrast is equally visible in this plane. MRI studies of rhotic articulation show that rhotics can involve coordinated tongue-tip, tongue-body, and labial gestures, with relevant structure extending beyond a single midsagittal view~\cite{proctor25_interspeech}.

Recent rtMRI resources provide synchronized audio, phonetic alignments, and articulatory contour annotations~\cite{narayanan2014,lim2021,shi2025}, enabling data-driven classification from vocal-tract motion. Prior contour- and image-based classification studies have shown that broad phonological categories are more accessible than fine-grained phone identity, but accurately predicting detailed phonetic labels from articulatory information alone remains difficult~\cite{saha18_interspeech,van2019cnn,pandey2021silent,park2026interpretable}. 
Related rtMRI work has explored both articulatory-to-acoustic mapping and acoustic-to-articulatory inversion, showing that speech acoustics and vocal-tract motion carry complementary information, while temporal alignment and speaker-specific variation remain important challenges~\cite{csapo20_interspeech,azzouz25_interspeech}. Acoustic--articulatory fusion has been applied to VCV classification from paired vocal-tract rtMRI and speech audio~\cite{yue24_interspeech}, and contrastive audio--MRI learning has been shown to improve phonological class recognition when audio is unavailable at inference time~\cite{liu2025tsd,ariasvergara24_interspeech}.
Beyond classification, synchronized speech has also been used as a cross-modal prior for vocal-tract MRI synthesis and reconstruction~\cite{perez2026speech,perez2026cross,shah2025mri2speech,hasan2026sirem}. Related work on vocal-tract segmentation further suggests that acoustic inputs can complement visual features when fused appropriately~\cite{liu2025vocseg}.

Self-supervised speech models (SSL) such as wav2vec~2.0, HuBERT, and WavLM provide strong general-purpose representations for downstream speech tasks~\cite{baevski2020,hsu2021,chen2022,yang2021}, including cross-lingual phoneme recognition~\cite{xu22b_interspeech}. Although probing and articulatory analyses show that these models encode phonetic, phonemic, and articulatory information~\cite{martin23_interspeech,shi2024direct}, their representations are not explicitly organized as interpretable phonological feature groups. For rtMRI analysis, a structured phonological level is useful because categories such as manner, place, voicing, and vowel quality are closely tied to articulatory organization~\cite{chomsky1968sound,browman1992articulatory}.

PhonoQ was introduced as a frame-level phonological feature recognizer for speech audio~\cite{arias2022analysis}, and PhonoQ~2.0 extends this framework with structured prediction heads for multilingual phonological modeling~\cite{hernandez2026}. Unless otherwise stated, \emph{PhonoQ} refers to PhonoQ~2.0 throughout this paper. PhonoQ provides the structured intermediate representation needed for our setting by predicting frame-level phonological features using an XLSR backbone fine-tuned for cross-lingual phoneme recognition~\cite{xu22b_interspeech}, a lightweight Conformer module~\cite{gulati20_interspeech}, and separate heads for different phonological feature groups. Because its representations are shaped by phonological supervision, PhonoQ offers a potential bridge between acoustic evidence, articulatory configuration, and phonological structure.

\begin{figure}[t]
\centering
\includegraphics[width=\columnwidth]{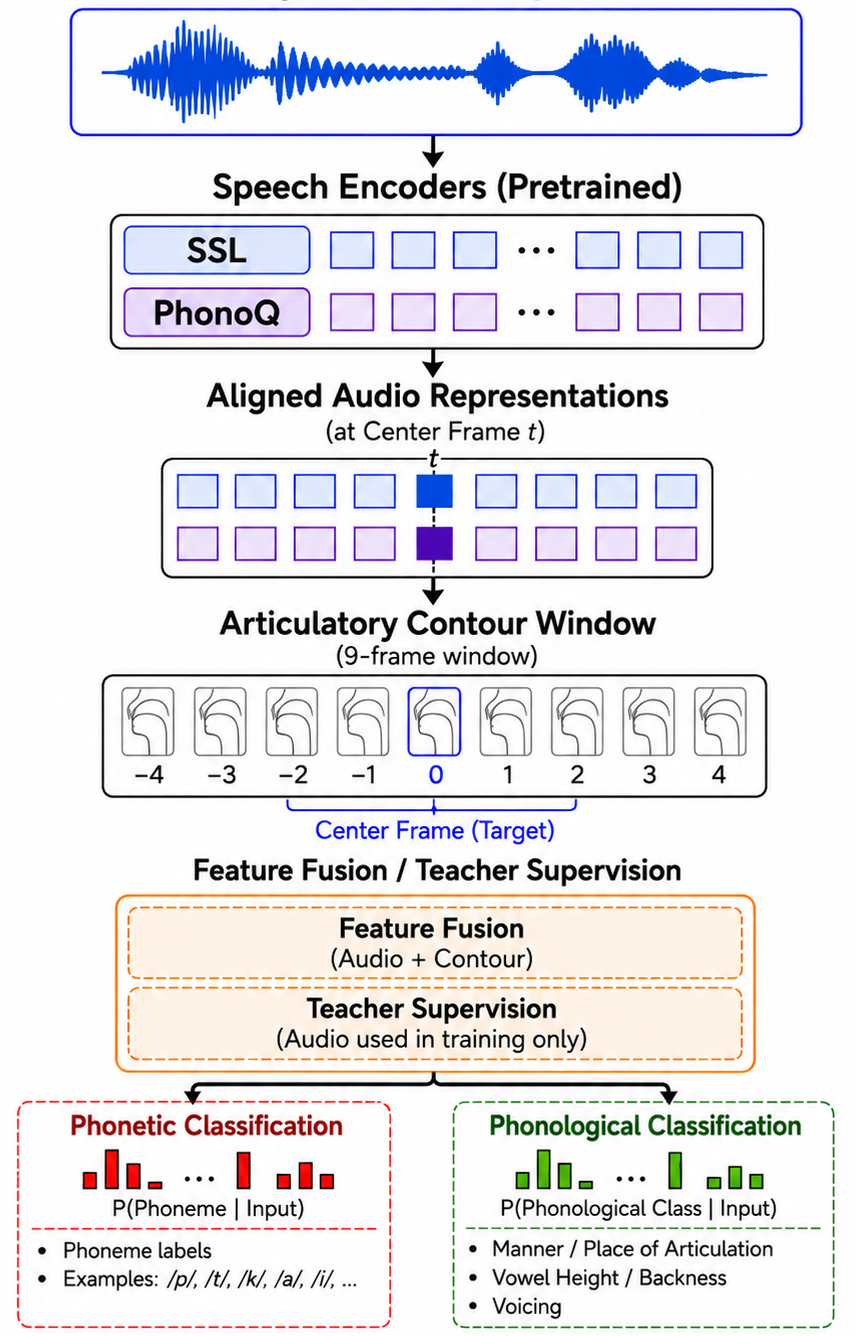}
\caption{\textbf{Method overview.} Audio-derived SSL and PhonoQ representations are aligned to articulatory contour windows and used either for feature fusion or as training-only teacher supervision.}
\label{fig:method_overview}
\end{figure}

This motivates a central question: can an audio model trained to predict structured phonological features provide useful information for articulatory speech modeling? Using the 75-Speaker Annot-16 dataset~\cite{shi2025}, we evaluate 39-way phoneme classification and five phonological targets: manner, consonant place, consonant voicing, vowel height, and vowel backness. We study PhonoQ in two settings: inference-time feature fusion and training-only teacher supervision for contour-only prediction. This separation allows us to distinguish cross-modal complementarity from the stricter question of whether audio-derived phonological information can be transferred to an articulatory model. Figure~\ref{fig:method_overview} summarizes the framework.

Our contributions are:
\begin{itemize}
\item A structured audio--articulatory framework that uses PhonoQ as a phonological representation source for rtMRI contour-based speech classification.
\item A systematic comparison of PhonoQ, HuBERT-large, and WavLM-large across 39-way phoneme classification and five phonological feature groups under unseen-speech and unseen-subject protocols.
\item An evaluation of two uses of synchronized audio: inference-time feature fusion and training-only teacher supervision for contour-only prediction.
\item A posterior-based analysis showing that PhonoQ exposes interpretable surface-sensitive patterns beyond canonical phone labels, including flapping-like \mbox{/t/}, \mbox{/t/--/r/} retraction or affrication, and nasal place assimilation.
\end{itemize}


\section{Experimental Setup} 
\label{sec:experimental_setup} 

\subsection{Dataset} 
\label{subsec:dataset} 

\begin{figure}[H]
\centering
\includegraphics[width=\columnwidth]{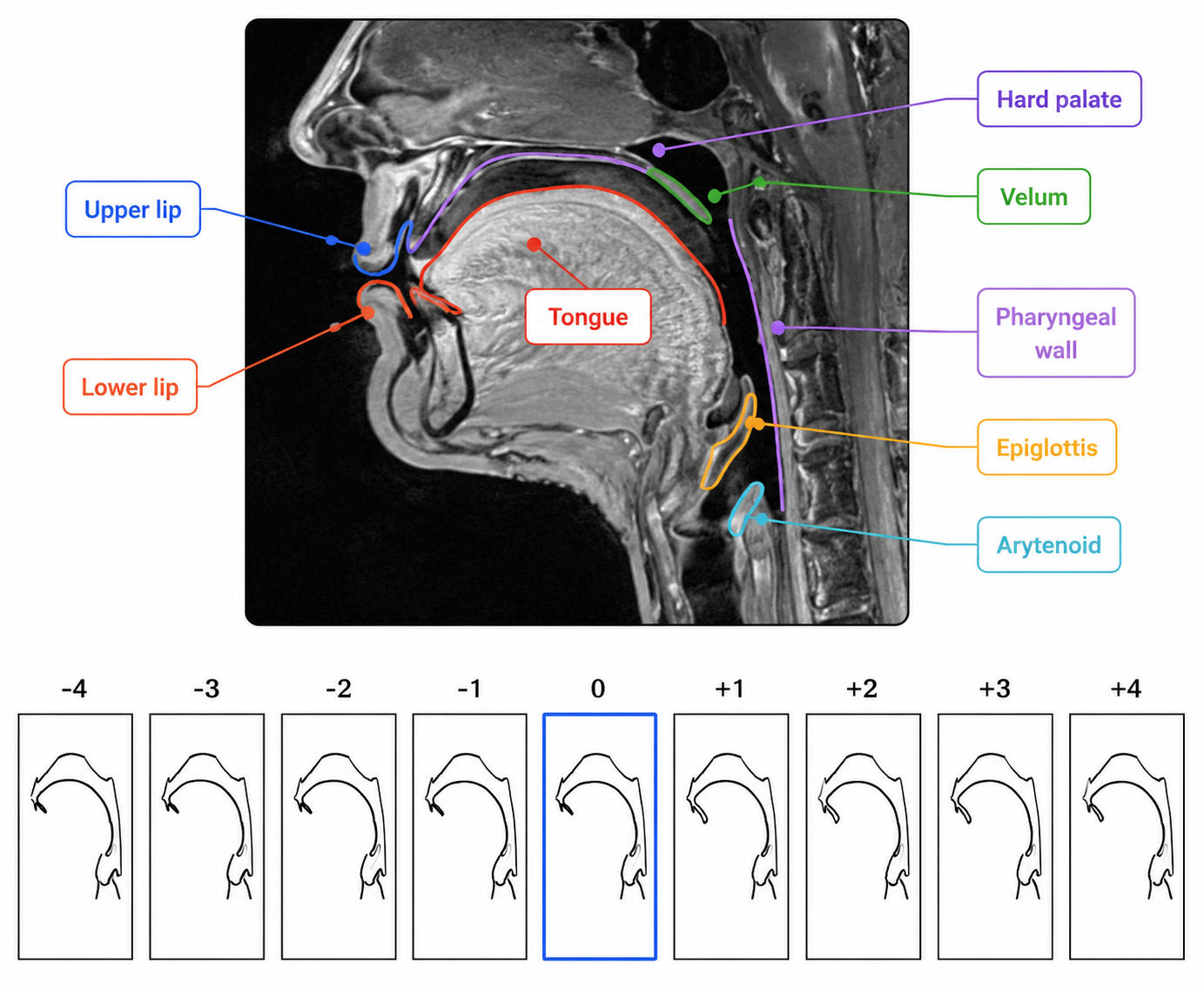}
\caption{\textbf{Articulatory contour input.} Annot-16 contour tracks represent vocal-tract shape as time-aligned articulatory vectors. Each example uses a 9-frame window centered on the target frame.}
\label{fig:contour_representation}
\end{figure}

We use the 75-Speaker Annot-16 dataset, an rtMRI speech production resource with synchronized speech audio, phonetic alignments, and expert-verified articulatory contour annotations from midsagittal vocal-tract MRI recordings~\cite{shi2025}. The contours trace multiple vocal-tract structures, including the tongue, lips, velum, pharyngeal wall, and hard palate. We use these contour tracks as a compact articulatory representation rather than using the raw rtMRI images directly. Each example is represented as a 9-frame contour window centered on the target frame, with each frame encoded as a 380-dimensional contour vector. Synchronized audio is used to extract audio-derived representations and teacher signals, but contour-only experiments use only articulatory contours at inference time. Figure~\ref{fig:contour_representation} illustrates the contour input.

\subsection{Prediction targets} 
\label{subsec:prediction_targets} 

We evaluate both fine-grained phonetic classification and structured phonological classification. The phonetic task is 39-way phone classification. The phonological tasks decompose phone identity into five interpretable feature groups: manner, consonant place, consonant voicing, vowel height, and vowel backness. This distinction tests whether articulatory contours are more predictive of broad phonological structure than of full phone identity, which may depend on acoustic distinctions weakly visible in midsagittal rtMRI.

\subsection{Audio-derived representations}
\label{subsec:audio_representations}

We extract audio-derived representations from three pretrained speech models: HuBERT-large, WavLM-large, and PhonoQ. HuBERT and WavLM serve as strong generic self-supervised speech representation baselines. In contrast, PhonoQ provides a phonologically structured representation because it is trained to predict frame-level phonological features from speech, including manner, place, voicing, vowel height, and vowel backness. Rather than using PhonoQ as a phoneme recognizer, we use its intermediate representations and posterior outputs as audio-derived phonological information.

All audio-derived features are extracted offline from synchronized waveforms resampled to 16 kHz and kept fixed during the downstream rtMRI experiments. For HuBERT-large and WavLM-large, we extract final-layer contextual hidden states from the speech encoder. For PhonoQ, we extract hidden representations from the Conformer module before the structured prediction heads. We also store the posterior tracks produced by the PhonoQ heads, which provide frame-level probabilities for the phonological feature groups.

Because the audio representations and articulatory contours operate at different frame rates, we align them by timestamp. For each contour example, we use the center-frame time to select nearby audio-model frames and pool them into a single audio-derived vector aligned with the contour window. The same procedure is used for HuBERT, WavLM, and PhonoQ. Continuous audio embeddings are standardized using training-split statistics, which are then applied to the development and test splits.

In the feature-fusion experiments, aligned audio-derived embeddings are available at inference time and are combined with contour representations. In the contour-only teacher-supervision experiments, audio-derived embeddings are used only during training; at inference time, the model receives only articulatory contours. For the posterior analysis, we use raw PhonoQ posterior tracks to examine how its phonological probabilities vary across selected phonetic contexts.

\subsection{Evaluation} 
\label{subsec:evaluation} 

We evaluate all systems under two complementary generalization protocols. In the \textit{unseen-speech} protocol, training and testing use different speech materials or elicitation conditions, while speakers are shared across splits. This setting tests whether the model generalizes across linguistic material and speaking context. In the \textit{unseen-subject} protocol, training and testing use disjoint speakers, which evaluates generalization to new vocal-tract anatomies and speaker-specific articulatory patterns. Model performance is reported primarily using macro-averaged F1, which gives equal weight to each class and is appropriate for the imbalanced phonetic and phonological label distributions.

\subsection{Training details}
\label{subsec:training_details}

All experiments use a filtered Annot-16 subset containing utterances with available phone-level alignments. After excluding the \texttt{bvt} and \texttt{vcv} task families and low-coverage speakers, the final subset contains 12 usable speakers and 16 speech tasks. In the unseen-subject protocol, speakers are disjoint across splits, with 8 speakers for training, 2 for development, and 2 for testing. In the unseen-speech protocol, the same speaker pool is retained, but speech materials are disjoint across splits: training uses the \texttt{grandfather} and \texttt{northwind} passages, \texttt{picture} description tasks, and \texttt{topic3}--\texttt{topic5}; development uses the \texttt{shibboleth} passage and \texttt{topic2}; and testing uses the held-out \texttt{rainbow} passage and \texttt{topic1}. Models are trained separately for each target and protocol using three random seeds.

Feature-fusion models use a two-layer bidirectional LSTM with hidden dimension 256 and gated fusion between contour and audio-derived representations. Audio features are projected to 512 dimensions before fusion. Models are trained for up to 50 epochs with early stopping after 8 epochs without development macro-F1 improvement, using AdamW with learning rate $10^{-3}$, weight decay $10^{-4}$, batch size 256, dropout 0.1, and no class weighting.

Contour-only teacher-supervision models use the same contour encoder but receive no audio input at inference time. Teacher-supervised systems add an instance-level contrastive loss between the contour representation and audio-derived teacher representations, with contrastive weight $\lambda=0.3$ and temperature $\tau=0.07$. The contour-only baseline is implemented by setting $\lambda=0$. These systems are trained for up to 40 epochs with early stopping after 6 epochs, using AdamW with learning rate $10^{-3}$, weight decay $10^{-4}$, batch size 256, dropout 0.2, and square-root inverse-frequency class weights computed from the training split.

\section{Modeling Framework}
\label{sec:modeling_framework}

\subsection{Feature-fusion classifier}
\label{subsec:feature_fusion_model}

We first evaluate whether PhonoQ-derived representations provide complementary information for audio--articulatory classification. In this setting, audio-derived features are available at inference time, so feature fusion is interpreted as an analysis of cross-modal complementarity rather than as a contour-only deployment setting.

For each target frame, the classifier receives a 9-frame aligned feature window. Depending on the system, this window contains contour features alone or contour features combined with one or more audio-derived representations. We compare six feature configurations: contour only, HuBERT, HuBERT+PhonoQ, WavLM, WavLM+HuBERT, and WavLM+PhonoQ.

All feature-fusion systems use the same sequence-classification backbone so that differences in performance can be attributed to the input representation rather than to changes in architecture. The contour sequence is encoded with a two-layer bidirectional LSTM. Audio-derived features are projected to a fixed dimension and combined with the contour representation using gated fusion. The center-frame representation is then passed to a target-specific linear classification head:
\begin{equation}
\hat{\mathbf{y}} = f_{\theta}(\mathbf{X}) .
\end{equation}
The model is trained with cross-entropy loss:
\begin{equation}
\mathcal{L}_{\mathrm{CE}}
= - \sum_{c=1}^{C} y_c \log \hat{y}_c .
\end{equation}
Here, $C$ is the number of classes for the target, $y_c$ is the one-hot reference label, and $\hat{y}_c$ is the predicted probability for class $c$. We train separate classifiers for each phonetic and phonological target.

\subsection{Contour-only teacher supervision}
\label{subsec:contour_teacher_model}

We next consider a stricter contour-only setting, where audio-derived information is available during training but not at inference time. The student model receives only the 9-frame articulatory contour window and uses the same contour encoder as the feature-fusion systems. This experiment tests whether information from synchronized audio can be transferred to an articulatory model that remains contour-only at test time.

The contour-only baseline is trained with cross-entropy loss. Teacher-supervised variants add an auxiliary instance-level contrastive loss that aligns the projected contour representation with the corresponding audio-derived teacher representation from the synchronized speech signal. We use a symmetric InfoNCE objective~\cite{oord2018representation} over paired contour--audio examples in each minibatch. The total training objective is
\begin{equation}
\mathcal{L}
=
\mathcal{L}_{\mathrm{CE}}
+
\lambda \mathcal{L}_{\mathrm{teacher}},
\end{equation}
where $\mathcal{L}_{\mathrm{CE}}$ is the classification loss, $\mathcal{L}_{\mathrm{teacher}}$ is the contrastive teacher-alignment loss, and $\lambda$ controls the strength of the teacher signal. The contour-only baseline corresponds to $\lambda=0$.

We evaluate HuBERT, WavLM, and PhonoQ as single teachers. We also evaluate weighted two-teacher supervision, where a generic SSL teacher is combined with PhonoQ by taking a weighted sum of their contrastive teacher-alignment losses. The mixing weight $\alpha$ controls the relative contribution of the SSL teacher and PhonoQ. In all cases, teacher representations are used only during training; at inference time, the model receives only articulatory contours.

\subsection{Posterior analysis}
\label{subsec:posterior_analysis_model}

In addition to using PhonoQ representations for classification and teacher supervision, we analyze PhonoQ posterior tracks to examine whether the model exposes interpretable surface-sensitive phonological cues. This analysis is not used to train the classifiers. Instead, it provides an interpretation of the phonological information available in the PhonoQ posterior space.

We extract canonical phone intervals from the phonetic alignments and match them to frame-level PhonoQ posterior predictions. For each interval, we average the relevant posterior dimensions over all frames contained in the interval:
\begin{equation}
    \bar{\mathbf{p}}_{i}
    =
    \frac{1}{|\mathcal{T}_{i}|}
    \sum_{t \in \mathcal{T}_{i}}
    \mathbf{p}_{t},
\end{equation}
where $\mathcal{T}_{i}$ is the set of PhonoQ frames aligned to phone interval $i$, and $\mathbf{p}_{t}$ is the PhonoQ posterior vector at frame $t$.

We summarize these interval-level posteriors by word and phonological context. Specifically, we examine canonical \mbox{/t/} intervals in likely flapping contexts, \mbox{/t/} before \mbox{/r/}, nasal intervals in assimilation-prone contexts, and matched canonical control words~\cite{stevens2000acoustic}. For each case, we report posterior dimensions relevant to the expected surface-sensitive pattern, such as stop, rhotic, affricate, alveolar, postalveolar, voiced, voiceless, vowel, and nasal posterior mass.

These analyses are intended as posterior summaries rather than manually verified allophonic labels. We therefore interpret the results as evidence consistent with surface-sensitive phonological patterns, not as automatic detection of flaps, glottal stops, or nasalized vowels.

\section{Results}
\label{sec:results}
We organize the results around the two roles of synchronized audio in our framework. First, we evaluate feature fusion to measure how much audio-derived representations complement articulatory contours when audio is available at inference time. Second, we evaluate contour-only teacher supervision to test whether audio-derived representations can improve an articulatory model without requiring audio at test time. We then analyze PhonoQ posterior tracks to examine whether the same structured representation exposes interpretable surface-sensitive phonological patterns.

\subsection{Feature-fusion results}
\label{subsec:results_fusion}

\begin{table*}[!t]
\centering
\caption{Test macro-F1 for phonetic and phonological classification. Contour is the articulatory-only baseline; other systems use audio-derived representations. Values are mean $\pm$ SD over three seeds.}
\label{tab:main_test_macro_f1}
\resizebox{\textwidth}{!}{%
\begin{tabular}{lllcccccc}
\toprule
Group & Protocol & Target & Contour & HuBERT & HuBERT+PhonoQ & WavLM & WavLM+HuBERT & WavLM+PhonoQ \\
\midrule
\multirow{2}{*}{Phonetic}
& Unseen speech & Phonemes
& 24.4 $\pm$ 0.3
& 56.0 $\pm$ 0.7
& 65.0 $\pm$ 0.2
& 64.0 $\pm$ 1.1
& 64.5 $\pm$ 0.6
& \textbf{69.0 $\pm$ 0.2} \\
& Unseen subject & Phonemes
& 19.3 $\pm$ 0.6
& 60.0 $\pm$ 0.4
& 64.1 $\pm$ 0.1
& 67.0 $\pm$ 0.3
& 67.5 $\pm$ 0.5
& \textbf{67.8 $\pm$ 0.1} \\
\midrule
\multirow{10}{*}{Phonological}
& \multirow{5}{*}{Unseen speech} & Cons. place
& 47.7 $\pm$ 1.0
& 72.4 $\pm$ 1.0
& 77.7 $\pm$ 0.7
& 79.8 $\pm$ 0.4
& 80.4 $\pm$ 0.1
& \textbf{82.4 $\pm$ 0.1} \\
& & Cons. voice
& 67.9 $\pm$ 1.1
& 85.5 $\pm$ 0.2
& 89.8 $\pm$ 0.2
& 89.3 $\pm$ 0.1
& 90.1 $\pm$ 0.3
& \textbf{92.1 $\pm$ 0.3} \\
& & Manner
& 38.1 $\pm$ 0.6
& 69.2 $\pm$ 0.9
& 73.8 $\pm$ 0.4
& 73.4 $\pm$ 0.5
& 73.9 $\pm$ 0.3
& \textbf{75.7 $\pm$ 0.2} \\
& & Vowel back.
& 65.8 $\pm$ 1.3
& 84.6 $\pm$ 0.2
& 88.2 $\pm$ 0.5
& 89.0 $\pm$ 0.2
& 89.5 $\pm$ 0.3
& \textbf{90.9 $\pm$ 0.2} \\
& & Vowel height
& 64.5 $\pm$ 0.2
& 82.4 $\pm$ 0.1
& 86.2 $\pm$ 0.1
& 86.3 $\pm$ 0.6
& 86.9 $\pm$ 0.3
& \textbf{88.5 $\pm$ 0.4} \\
\cmidrule{2-9}
& \multirow{5}{*}{Unseen subject} & Cons. place
& 41.1 $\pm$ 1.0
& 72.6 $\pm$ 0.9
& 75.2 $\pm$ 0.8
& 79.5 $\pm$ 0.1
& \textbf{80.1 $\pm$ 0.5}
& 79.9 $\pm$ 0.6 \\
& & Cons. voice
& 61.2 $\pm$ 1.0
& 84.9 $\pm$ 0.8
& 86.7 $\pm$ 0.3
& 89.6 $\pm$ 0.4
& 89.5 $\pm$ 0.6
& \textbf{90.1 $\pm$ 0.0} \\
& & Manner
& 32.1 $\pm$ 0.3
& 66.6 $\pm$ 0.8
& 70.1 $\pm$ 0.2
& 71.9 $\pm$ 0.9
& 71.9 $\pm$ 0.6
& \textbf{72.6 $\pm$ 0.2} \\
& & Vowel back.
& 60.9 $\pm$ 0.7
& 85.2 $\pm$ 0.2
& 87.0 $\pm$ 0.4
& 89.4 $\pm$ 0.3
& \textbf{89.7 $\pm$ 0.2}
& 89.6 $\pm$ 0.4 \\
& & Vowel height
& 58.7 $\pm$ 2.0
& 84.4 $\pm$ 0.5
& 86.4 $\pm$ 0.4
& 89.3 $\pm$ 0.3
& \textbf{89.7 $\pm$ 0.4}
& 89.2 $\pm$ 0.1 \\
\bottomrule
\end{tabular}%
}
\end{table*}

Table~\ref{tab:main_test_macro_f1} reports test macro-F1 for the feature-fusion setting, where audio-derived representations are available at inference time. These results therefore measure audio--articulatory complementarity rather than contour-only performance.

The contour-only baseline is substantially below all audio-fusion systems across both phonetic and phonological targets, indicating that synchronized audio representations provide information not fully captured by articulatory contours. For 39-way phoneme classification, WavLM+PhonoQ gives the best results in both protocols, reaching 69.0\% macro-F1 for unseen speech and 67.8\% for unseen subjects. This improves over WavLM alone, which reaches 64.0\% and 67.0\%, respectively. A similar trend is observed for HuBERT, where adding PhonoQ increases macro-F1 from 56.0\% to 65.0\% in the unseen-speech setting and from 60.0\% to 64.1\% in the unseen-subject setting.

For phonological classification, WavLM+PhonoQ is strongest in the unseen-speech protocol, obtaining the best result for all five targets. The pattern is more mixed under unseen-subject generalization: WavLM+PhonoQ remains best for consonant voicing and manner, while WavLM+HuBERT is slightly stronger for consonant place, vowel backness, and vowel height. Thus, PhonoQ is most consistently beneficial in the unseen-speech setting, while the best feature combination under speaker generalization depends on the target.\

Averaged across protocols, WavLM+PhonoQ is the strongest overall feature-fusion system, reaching 68.4\% macro-F1 for phonetic classification and 85.1\% for phonological classification. The gains are especially clear when PhonoQ is added to HuBERT, increasing the average phonetic score from 58.0\% to 64.6\% and the average phonological score from 80.5\% to 83.8\%. Overall, these results suggest that PhonoQ does not replace generic SSL representations, but contributes complementary structured phonological information when combined with HuBERT or WavLM.

\subsection{Audio-teacher supervision}
\label{subsec:results_contour_contrastive}

We next evaluate contour-only models trained with audio-derived teacher supervision. Table~\ref{tab:contour_teacher_overall} reports test macro-F1 separately for the unseen-speech and unseen-subject protocols, averaged over five phonological targets and three seeds.

All teacher-supervised systems improve over the contour-only baseline. In the unseen-speech protocol, single-teacher systems reach 59.2--59.3\% macro-F1, compared with 57.4\% for contour only. The best result is obtained by HuBERT+PhonoQ at 59.5\%, closely followed by WavLM+PhonoQ at 59.4\%. The same trend holds for unseen subjects: contour only reaches 51.8\%, single-teacher systems reach 53.4--53.7\%, and both two-teacher systems reach 54.1\%. These gains are modest but consistent, indicating that audio-derived supervision can regularize contour-only phonological prediction without requiring audio at inference time.

\begin{table}[t]
\centering
\caption{Contour-only inference with audio-derived teacher supervision. Values are test macro-F1 averaged over five phonological targets and three seeds within each protocol.}
\label{tab:contour_teacher_overall}
\setlength{\tabcolsep}{5pt}
\resizebox{\columnwidth}{!}{%
\begin{tabular}{lcccc}
\toprule
\multirow{2}{*}{System} & \multicolumn{2}{c}{Unseen speech} & \multicolumn{2}{c}{Unseen subject} \\
\cmidrule(lr){2-3}\cmidrule(lr){4-5}
& F1 & $\Delta$ & F1 & $\Delta$ \\
\midrule
Contour-only & 57.4 & -- & 51.8 & -- \\
HuBERT & 59.2 & +1.8 & 53.4 & +1.5 \\
PhonoQ & 59.2 & +1.8 & 53.6 & +1.7 \\
WavLM & 59.3 & +1.8 & 53.7 & +1.9 \\
HuBERT+PhonoQ & \textbf{59.5} & \textbf{+2.1} & \textbf{54.1} & \textbf{+2.3} \\
WavLM+PhonoQ & 59.4 & +2.0 & \textbf{54.1} & \textbf{+2.3} \\
\bottomrule
\end{tabular}%
}
\end{table}

The gains are most visible for lower-baseline targets. For phone manner, PhonoQ improves over contour only from 40.4\% to 42.9\% in the unseen-speech setting and from 34.0\% to 35.6\% in the unseen-subject setting. For consonant place under unseen-subject generalization, contour only reaches 41.2\%, while WavLM and HuBERT+PhonoQ reach 44.4\% and 44.5\%, respectively. These target-level results suggest that audio-derived teacher supervision is particularly helpful when contour-only prediction is difficult.

Overall, the contour-only results complement the feature-fusion findings. While feature fusion shows that PhonoQ adds information when audio-derived features are available at inference time, the teacher-supervision setting shows that PhonoQ can also provide useful training-time supervision for an articulatory model that remains contour-only at test time.

\subsection{PhonoQ posterior analysis}
\label{subsec:results_posterior}

\begin{figure}
    \centering
    \includegraphics[width=\columnwidth]{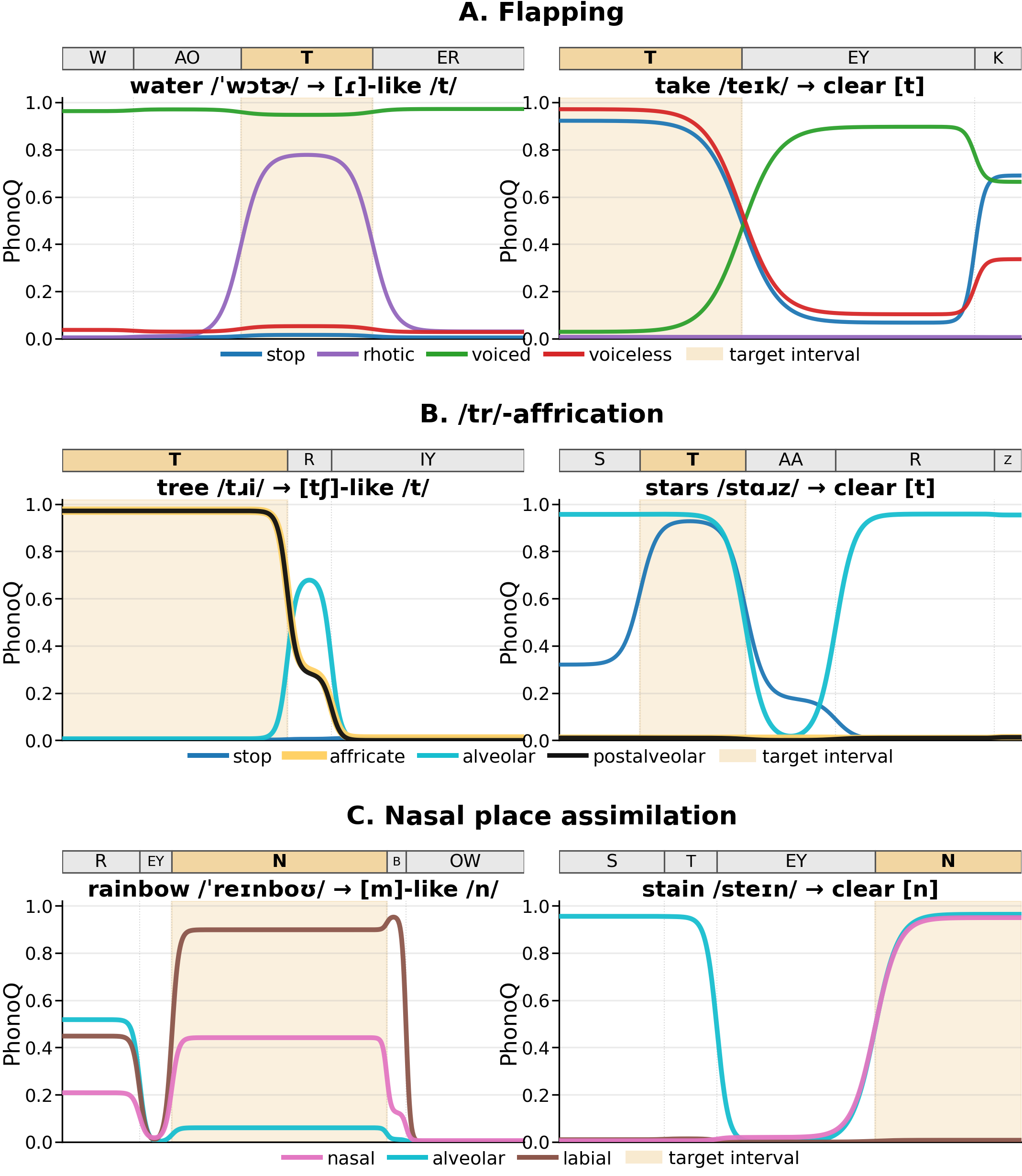}
    \caption{PhonoQ posterior tracks for three surface-sensitive contexts
paired with canonical controls: \textbf{(A)}~flapping (\textit{water}
vs.\ \textit{take}), \textbf{(B)}~/tr/-affrication (\textit{tree}
vs.\ \textit{stars}), and \textbf{(C)}~nasal place assimilation
(\textit{rainbow} vs.\ \textit{stain}). Shaded regions mark the
target phone interval.}
    \label{fig:phonoq-textgrid}
\end{figure}

Figure~\ref{fig:phonoq-textgrid} illustrates PhonoQ posterior behavior in selected surface-sensitive phonological contexts, paired with canonical controls. This analysis examines whether the PhonoQ posterior space exposes interpretable phonological cues beyond the canonical phone labels.

For canonical /t/ intervals in likely flapping contexts, the posterior distributions show lower stop and voiceless mass and higher voiced and rhotic mass than clear /t/ controls. For example, \textit{water}, \textit{pretty}, \textit{little}, and \textit{better} show substantially higher voiced posterior mass than clear /t/ controls such as \textit{two}, \textit{take}, \textit{still}, and \textit{started}. This pattern is consistent with flapping-like surface realizations.

A second pattern appears for /t/ before /r/. Words such as \textit{street} and \textit{tree} show high affricate and postalveolar posterior mass, whereas clear /t/ controls such as \textit{stars} and \textit{details} remain strongly stop-like and alveolar. This suggests that PhonoQ posteriors capture retraction- or affrication-like cues in /t/-/r/ contexts.

Finally, the nasal place assimilation example shows that \textit{rainbow}, where canonical /n/ precedes /b/, has increased labial posterior mass and reduced alveolar mass during the target nasal interval relative to the clear /n/ control \textit{stain}. This pattern is consistent with anticipatory labial assimilation.

Overall, these posterior summaries suggest that PhonoQ provides interpretable surface-sensitive phonological evidence that is not explicitly encoded by canonical phone labels alone~\cite{browman1992articulatory}. This supports its use not only as an audio-derived representation for classification, but also as a linguistically structured source of supervision and analysis.

\section{Discussion and Conclusion}
\label{sec:discussion_conclusion}

This work examined whether structured phonological representations from PhonoQ are useful for audio--articulatory rtMRI speech classification. The strongest evidence comes from the feature-fusion setting, where PhonoQ representations improve performance when combined with generic SSL features from HuBERT or WavLM. The best overall system, WavLM+PhonoQ, achieves the highest average performance for both 39-way phoneme classification and structured phonological classification, although WavLM+HuBERT remains competitive for some unseen-subject phonological targets. These results suggest that PhonoQ provides information that is complementary to SSL speech models.

The training-only teacher-supervision results provide a stricter test of whether audio-derived information can be transferred to articulatory models. In this setting, audio is used only during training, and the model receives only contour features at inference time. The gains are smaller than in feature fusion, but they are consistent across both generalization protocols. This indicates that audio-derived teacher signals can regularize contour-only phonological prediction, although much of the acoustic information available to SSL and PhonoQ models is not fully recoverable from midsagittal articulatory contours alone. The strongest interpretation is therefore not that PhonoQ solves contour-only phonological recognition, but that structured audio-derived supervision provides a useful training signal for difficult phonological targets.

The posterior analysis further illustrates why PhonoQ may be useful in this setting. Because PhonoQ predicts structured phonological features rather than only phone identities, its posterior tracks can expose interpretable surface-sensitive cues. The examples in Figure~\ref{fig:phonoq-textgrid} show posterior patterns consistent with flapping-like /t/ realizations, /t/-/r/ affrication or retraction, and nasal place assimilation. In the nasal assimilation example, \textit{rainbow}, where canonical /n/ precedes /b/, shows increased labial posterior mass and reduced alveolar mass relative to the clear /n/ control \textit{stain}. These analyses are not manually verified allophonic annotations, but they show that PhonoQ can provide a linguistically meaningful view of speech variation beyond canonical phone labels.

Several limitations remain. First, the feature-fusion experiments use audio-derived representations at inference time and should therefore be interpreted as an analysis of cross-modal complementarity, not as an MRI-only deployment scenario. Second, the contour-only improvements are modest, suggesting that stronger articulatory encoders, better temporal modeling, or more targeted teacher objectives may be needed to transfer more phonological information from audio to contours. Third, the posterior analysis is illustrative and should be extended with larger-scale quantitative analyses and, where possible, manual phonetic validation of surface forms.

Overall, the results support the use of structured phonological representations as an intermediate level between low-level articulatory motion and fine-grained phone identity. PhonoQ-derived features improve audio--articulatory classification when used in feature fusion, provide modest but consistent benefits as training-time teacher signals for contour-only inference, and offer interpretable posterior evidence for surface-sensitive phonological patterns. These findings suggest that phonologically structured audio models can be useful tools for rtMRI speech analysis, especially when the goal is not only classification accuracy but also interpretable links between articulation, acoustics, and phonological structure.

\section{Generative AI Use Disclosure}
Generative artificial intelligence tools were used to assist with language editing, clarity of presentation, and the preparation of illustrative visual material. Any AI-assisted image content was manually reviewed and verified by the authors for consistency with the intended anatomical and methodological depiction. No AI-generated content was used to produce experimental results, labels, quantitative analyses, or scientific interpretations. All research ideas, methodology, experiments, results, and interpretations were conceived, conducted, and validated by the authors, who take full responsibility for the originality, validity, and integrity of the work.

\bibliographystyle{IEEEtran}
\bibliography{refs}

\end{document}